\documentclass[dvipsnames,format=sigconf,anonymous=false,review=false]{acmart} 

\def\vec#1{\mathchoice{\mbox{\boldmath$\displaystyle#1$}}
{\mbox{\boldmath$\textstyle#1$}}
{\mbox{\boldmath$\scriptstyle#1$}}
{\mbox{\boldmath$\scriptscriptstyle#1$}}}
\usepackage{booktabs} % For formal tables
\usepackage{amsmath}
\usepackage{algorithm}
\usepackage[noend]{algpseudocode}
\usepackage[T1]{fontenc}
\usepackage{tabularx}
\usepackage{multirow}
\usepackage{array}
\usepackage{pbox}

\usepackage{color,soul}
\usepackage{nameref}

\usepackage{pgfplots}
\pgfplotsset{compat=1.18}
\usepackage{graphicx}
\usepackage{subcaption}
\usepackage{tikz}
\usetikzlibrary{intersections, pgfplots.fillbetween}
\usepackage{siunitx}

\usepackage{url}
\usepackage{bm}

\definecolor{darkgreen}{rgb}{0.30, 0.50, 0.0}

\makeatletter
\renewcommand{\ALG@name}{Pseudocode}
\makeatother

\newcolumntype{L}[1]{>{\raggedright\let\newline\\\arraybackslash\hspace{0pt}}m{#1}}
\newcolumntype{C}[1]{>{\centering\let\newline\\\arraybackslash\hspace{0pt}}m{#1}}
\newcolumntype{R}[1]{>{\raggedleft\let\newline\\\arraybackslash\hspace{0pt}}m{#1}}

\graphicspath{{images/}}

\copyrightyear{2026}
\acmYear{2026}
\setcopyright{cc}
\setcctype{by-nc-nd}
\acmConference[GECCO '26]{Genetic and Evolutionary Computation Conference}{July 13--17, 2026}{San Jose, Costa Rica}
\acmBooktitle{Genetic and Evolutionary Computation Conference (GECCO '26), July 13--17, 2026, San Jose, Costa Rica}
\acmDOI{10.1145/3795095.3805173}
\acmISBN{979-8-4007-2487-9/2026/07}

 \hypersetup{draft}
 
\begin{document}
 	\title[The hop-like problem nature]{The hop-like problem nature -- unveiling and modelling new features of real-world problems}

 	\author{Michal W. Przewozniczek}
 	\affiliation{
 		% \institution{Dep. of Systems and Comp. Networks}
 		\institution{Wroclaw Univ. of Science and Techn.}
 		\city{Wroclaw}
 		\country{Poland} 
 	}
 	\email{michal.przewozniczek@pwr.edu.pl}

         \author{Bartosz Frej}
         \affiliation{
         	% \institution{Fac. of Pure and Applied Mathematics}
         	\institution{Wroclaw Univ. of Science and Techn.}
         	\city{Wroclaw}
         	\country{Poland} 
         }
         \email{bartosz.frej@pwr.edu.pl}

        \author{Marcin M. Komarnicki}
 	\affiliation{
 		% \institution{Dep. of Computational Intelligence}
 		\institution{Wroclaw Univ. of Science and Techn.}
 		\city{Wroclaw}
 		\country{Poland} 
 	}
 	\email{marcin.komarnicki@pwr.edu.pl}

	\renewcommand{\shortauthors}{}

	\begin{abstract}
      Benchmarks are essential tools for the optimizer's development. Using them, we can check for what kind of problems a given optimizer is effective or not. Since the objective of the Evolutionary Computation field is to support the tools to solve hard, real-world problems, the benchmarks that resemble their features seem particularly valuable. Therefore, we propose a hop-based analysis of the optimization process. We apply this analysis to the NP-hard, large-scale real-world problem. Its results indicate the existence of some of the features of the well-known Leading Ones Problem. To model these features well, we propose the Leading Blocks Problem (LBP), which is more general than Leading Ones and some of the benchmarks inspired by this problem. LBP allows of the assembly of new types of hard optimization problems that are not handled well by the considered state-of-the-art Genetic Algorithm (GA). Finally, the experiments reveal what kind of mechanisms must be proposed to improve GAs' effectiveness while solving LBP and the considered real-world problem.
    \end{abstract}
	
% 	%
% 	% The code below should be generated by the tool at
% 	% http://dl.acm.org/ccs.cfm
% 	% Please copy and paste the code instead of the example below. 
% 	%
% 	\begin{CCSXML}
% 		<ccs2012>
% 		<concept>
% 		<concept_id>10010147.10010178</concept_id>
% 		<concept_desc>Computing methodologies~Artificial intelligence</concept_desc>
% 		<concept_significance>500</concept_significance>
% 		</concept>
% 		</ccs2012>
% 	\end{CCSXML}
	
% 	\ccsdesc[500]{Computing methodologies~Artificial intelligence}


	

	 \maketitle

\section{Introduction}
Genetic Algorithms (GAs) are highly effective optimizers in solving many hard, real-world problems \cite{whitley2019}. A given mechanism may improve GA's effectiveness for one problem and deteriorate for another. Therefore, some GA-related research focuses on identifying those problem features that seem to influence the GAs' performance. On this base, benchmark problems are proposed that share (or do not share) particular features observed based on real-world problems. Important examples of such benchmark tools may be the deceptive functions \cite{decFunc,decBimodalOld} and the Hierarchical-If-And-Only-If problems \cite{hiff,hiff2}.\par

In this work, we focus on unveiling the features of an NP-hard real-world problem. To this end, we propose the hop-based analysis. We propose the Leading Blocks Problem (LBP) to model the observed features. We show that our proposition is more general than the benchmark problems arising from the idea of the Leading Ones Problem \cite{leadingOnesDoer,SoLeadingOnes,Lob1,Lob2}. Moreover, the results obtained for state-of-the-art GAs indicate what kind of problem decomposition techniques seem necessary to solve LBP (and, consequently, the considered real-world problem) effectively.\par

The rest of this paper is organized as follows. In the next section, we present the definition of the considered real-world problem, the optimizers dedicated to solving it, and the fundamental analysis of its features. Section \ref{sec:RelatedWork:leadingOnes} presents the Leading Ones Problem and its versions as well as their similarity to the considered real-world problem. In the fourth section, we define LBP, propose the hop-based analysis and use it to present the features of LBP. Sections \ref{sec:expWPlfl}-\ref{sec:leadingBlocks} present the results and the analysis of the performed experiments. Finally, the last section concludes this work and identifies the most important future research directions.

\section{WP\_LFL Problem}
\label{sec:wpLflf}

The optimization of backbone computer networks is an important real-world problem \cite{PioroMedhi}. In this paper, we consider the flow allocation according to the given network topology and the capacity of network links \cite{PioroMedhi}. 
The objective of the considered flow assignment problem is to set the communication channels between a given set of network node pairs (demands) according to the link capacity constraint for the given network topology. A solution is encoded as a vector of the length equal to the number of demands. Each variable indicates which route should be used to set a communication channel between a given demand. As a quality measure, we employ the Lost Flow in Link (LFL) function \cite{walkowiakLFL}. The considered problem is NP-complete \cite{mupActive} and denoted as WP\_LFL \cite{walkowiakLFL}.\par

\subsection{Problem definition}
\label{sec:wpLflf:def}

The input information consists of network topology and a set of demands. Each demand is an amount of information to be sent between the given start and end nodes. To fulfil a demand, we must set a route in the network. Thus, a solution to the WP\_LFL problem is a list of routes that satisfy all demands. The solution is feasible if the link capacity constraint is not violated. WP\_LFL optimization improves the preparation of the network for the link breakdown scenario. We consider the same WP\_LFL formulation, presented in Table~\ref{tab:wpLFL}, and instances as in \cite{mupActive,nonBinaryELL}. They consist of 1260 and 2500 demands, which indicates using the same number of genes. Thus, we can state that, in the case of WP\_LFL, we consider the large-scale global optimization problem \cite{lsgoComplex}.

\begin{table}[]
 \caption{WP\_LFL defintion}
	\label{tab:wpLFL}
 % \scriptsize
\scriptsize
\begin{tabular}{l|l}
\hline
\multicolumn{2}{c}{\textit{Sets}}                  \\
\hline
$V$ - set of $n$ vertices (network nodes) & $A$ - set of $m$ arcs (directed network links) \\
$P$ - set of $q$ network connections  & $\Pi_p$ - the index set of routes for connection $p$ \\
\multicolumn{2}{l}{$X_r$ - set of variables $x^k_p$. $X_r$ determines the unique set of currently selected working paths}\\
\hline
\multicolumn{2}{c}{\textit{Indices}}                  \\
\hline
$p$ - subscript that refers to demands & $k$ - subscript that refers to candidate routes \\
$a$ - subscript that refers to arcs &  $r$ - subscript that refers to selections \\
\hline
\multicolumn{2}{c}{\textit{Other}}                  \\
\hline
$o(a)$ - the start node of arc $a$ & $d(a)$ - the end node of arc $a$\\
\hline
\multicolumn{2}{c}{\textit{Constants}}                  \\
\hline
\multicolumn{2}{l}{$\delta^k_{pa}$ - if arc $a$ is a part of path $k$ that is a part of connection $p$ it equals 1; otherwise it equals 0}\\
$Q_p$ - the volume of connection $p$ & $c_a$ - capacity of arc $a$\\
\hline
\multicolumn{2}{c}{\textit{Variables}}                  \\
\hline
\multicolumn{2}{l}{$x^k_p$ - equals 1 if working route $k\in \Pi_p$ is a part of connection $p$, otherwise it equals 0}\\
$g^{in}_v=\sum_{a \in A:d(a)=v}f_a$ & $e^{in}_v=\sum_{a \in A:d(a)=v}c_a$ \\
$g^{out}_v=\sum_{a \in A:o(a)=v}f_a$ & $e^{out}_v=\sum_{a \in A:o(a)=v}c_a$

\end{tabular}
\end{table}

%\textit{Sets}\\
%$V$ - set of $n$ vertices (the network nodes)\\
%$A$ - set of $m$ arcs (the directed links of the network)\\
%$P$ - set of $q$ network connections\\
%$\Pi_p$ -	the index set of routes for connection $p$\\
%$X_r$ - set (selection) of variables $x^k_p$. $X_r$ determines the unique set of currently selected working paths\\
%\textit{Indices}\\
%$p$ - subscript that refers to connections (demands)\\
%$k$ - subscript that refers to candidate routes\\
%$a$ - subscript that refers to arcs\\
%$r$ - subscript that refers to selections\\
%\textit{Other}\\
%$o(a)$ - the start node of arc $a$\\
%$d(a)$ - the end node of arc $a$\\
%\textit{Constants}\\
%$\delta^k_{pa}$ - if arc $a$ is a part of path $k$ that is a part of connection $p$ it equals 1; otherwise it equals 0\\
%$Q_p$ - the bandwidth requirement (volume) of connection $p$\\
%$c_a$ - capacity of arc $a$\\
%\textit{Variables}\\
%$x^k_p$ - decision variable (equals 1 if working route $k\in \Pi_p$ is a part of connection $p$, otherwise it equals 0)\\
%$f_a$ - flow of arc a\\
%$g^{in}_v=\sum_{a \in A:d(a)=v}f_a$ - the summarized flow of arcs that enter $v$;\\
%$e^{in}_v=\sum_{a \in A:d(a)=v}c_a$ - the summarized capacity of arcs that enter $v$;\\
%$g^{out}_v=\sum_{a \in A:o(a)=v}f_a$ - the summarized flow of arcs that leave $v$;\\
%$e^{out}_v=\sum_{a \in A:o(a)=v}c_a$ - the summarized capacity of arcs that leave $v$;\par

The $o:A \rightarrow V$ and $d:A \rightarrow V$  functions denote the origin and destination node of each arc. For each $a\in A$ the set of incoming arcs of $d(a)$ except $a$ is defined as $in(a)=\{i\in A| d(i)=d(a), i\neq a \}$, and the set of outgoing arcs of $o(a)$ except $a$ is defined as $out(a)=\{i\in A| o(i)=o(a), i\neq a \}$.\par

\textbf{Definition 1.} \textit{The global non-bifurcated multi commodity flow denoted by $\underline{f}=[f_1,f_2,...,f_m]$ is defined as a vector of flows in all arcs. The flow $\underline{f}$ is feasible if for every arc $a\in A$ the following inequality holds $\forall a \in A: f_a \leq c_a$. This inequality is a capacity constraint. It guarantees that the flow does not exceed the capacity in all arcs of the network.}

%For the sake of simplicity, we use the following function \cite{walkowiakLFL}: $\epsilon(x)= \begin{cases}	0, & x \leq 0\\	x, & x > 0	\end{cases}$

%\begin{equation}
%	\label{eq:def02}
%	\epsilon(x)=
%	\begin{cases}
%		0 & \text{for } x \leq 0\\
%		x & \text{for } x > 0
%	\end{cases}
%\end{equation}

%\begin{equation}
%	\label{eq:def02}
%	\epsilon(x)=
%	\begin{cases}
%		0, & x \leq 0\\
%		x, & x > 0
%	\end{cases}
%\end{equation}

In WP\_LFL, we consider the link, i.e., $b\in A$, failure scenario. The origin node of arc $b$ must reroute the flow to perform a local repair. This requirement makes the residual capacity of all arcs except $b$ that leave node $o(b)$ the potential bottleneck. Therefore, if 
\begin{equation}
    \small
    f_b \leq \sum_{a \in out(b)} (c_a - f_a),
\end{equation} then the residual capacity of links other than $b$ that leave node $o(b)$ will be sufficient to restore the flow. Nevertheless, if 
\begin{equation}
    \small
    f_b > \sum_{a \in out(b)} (c_a - f_a),
\end{equation}
then the residual capacity of other links that leave node $o(b)$ will not be sufficient, and some of the flow that was using link $b$ will be lost. We define the $LA^{out}$ function (using the definitions of $g^{out}_{o(b)}$, $e^{out}_{o(b)}$ and ineaqualities defined above), to define the flow that will be lost:
\begin{equation}
    \small
    LA^{out}_b(\underline{f}) = \epsilon(g^{out}_{o(b)} - (e^{out}_{o(b)} - c_b) ),
\end{equation}
where
\begin{equation}
    \small
    \epsilon(x)= \begin{cases}	0, & x \leq 0\\	x, & x > 0	\end{cases}.
\end{equation}
Using this formula, we can define the flow that is lost for all arcs that leave node $v$:
\begin{equation}
    \small
    LN^{out}_v(\underline{f}) = \sum_{a:o(a)=v} \epsilon(g^{out}_{v} - (e^{out}_{v} - c_a) ) = \sum_{a:o(a)=v} LA^{out}_v(\underline{f}).
\end{equation}
We also define function $LN^{in}_v(\underline{f})$ that computes the amount of flow that is lost for the arcs that enter node $v$. We use $LN^{out}_v(\underline{f})$ and $LN^{in}_v(\underline{f})$, to measure the level of a network preparation for the scenario of link breakdown: 
\begin{equation}
    \small
    LFL(\underline{f}) = \frac{\sum_{v \in V} (LN^{in}_v(\underline{f}) + LN^{out}_v(\underline{f}))}{2}.
\end{equation}\par

%More information about LFL may be found in \cite{walkowiakLFL}. WP\_LFL is defined as $\min_{\underline{f}} \quad LFL(\underline{f})$

The definition of the WP\_LFL optimization problem \cite{mupActive,nonBinaryELL} is as follows.

\begin{equation}
    \small
	\label{eq:def08}
	\min_{\underline{f}} \quad LFL(\underline{f}),
\end{equation}
subject to
\begin{equation}
    \small
    \label{eq:c1}
    \sum_{k \in \Pi_p}x^k_p = 1 \quad  \forall p \in P,
\end{equation}
\begin{equation}
    \small
    \label{eq:c2}
    x^k_p \in \{0,1\} \quad \forall p \in P, \forall k \in \Pi_p,
\end{equation}
\begin{equation}
    \small
    \label{eq:c3}
    f_a = \sum_{p \in P} \sum_{k \in \Pi_p} \delta^k_{pa} x^k_p Q_p \quad \forall a \in A,
\end{equation}
\begin{equation}
    \small
    \label{eq:c4}
    f_a \leq c_a \quad \forall a \in A,
\end{equation}
where (\ref{eq:c1}) and (\ref{eq:c2}) indicate that a single route is used for each communication connection, (\ref{eq:c3}) is the link flow definition, and (\ref{eq:c4}) is the link capacity constraint. Together with (\ref{eq:def08}), these formulas define the considered problem as the 0/1 NP problem with linear constraints.

% $\sum_{k \in \Pi_p}x^k_p = 1 \quad  \forall p \in P$ (a single route is used for each communication connection), $f_a = \sum_{p \in P} \sum_{k \in \Pi_p} \delta^k_{pa} x^k_p Q_p \quad \forall a \in A$ (link flow definition), $f_a \leq c_a \quad \forall a \in A$ (link capacity constraint), and $x^k_p \in \{0,1\} \quad \forall p \in P, \forall k \in \Pi_p$, which define the considered problem as the 0/1 NP problem with linear constraints.

%More information about WP\_LFL, including the analysis of its features from the optimization point of view, examples, and explanations of why linkage learning might be crucial to solving this problem effectively can be found in \cite{nonBinaryELL}.\par

%LSGO problems are considered hard not only because of their NP-hard or NP-complete nature but also due to their size \textcolor{red}{[BIB]}.\par

%. Thus, 1260 or 2500 genes are necessary to encode the full solution. 
%As shown in \cite{lldsi}, the number of encodable solutions for the considered test cases is in the range from $10^{1052}$ up to $10^{7671}$. Thus, we can state that, in case of WP\_LFL, we consider the Large-Scale Global Optimization problem (LSGO) \textcolor{red}{[BIB LSGOsize]}. Note that LSGO problems are considered as particularly hard not only because of its NP-hard or NP-complete nature but also due to their size \textcolor{red}{[BIB]}.\par

\subsection{Problem-dedicated optimizers}
\label{sec:wpLflf:opt}
    WP\_LFL is a discrete problem in which, for each gene, many values are available \cite{nonBinaryELL}. The group of WP\_LFL-dedicated optimizers includes the exact methods \cite{Walkowiak_iccs03}, the Lagrangian Relaxation Heuristic (LRH) \cite{Dias_icss03} hybridized with the Flow Deviation for Primary Routes algorithm \cite{FDfratta}, and standard GAs \cite{hefan22}.\par

    Multi Population Pattern Searching Algorithm (MuPPetS) \cite{muppets} is a base for some of the effective WP\_LFL-dedicated optimizers. MuPPetS uses the idea of messy coding \cite{messyGA}, where each individual may encode only a part of the whole genotype. A single individual encoding a complete genotype (denoted as \textit{Competitive Template}, CT) is maintained to rate messy-coded individuals. The genes from CT supplement the missing genes in the given messy-coded individual. Thus, a messy-coded individual can be considered as a modification of the CT genotype (the genes specified by the messy-coded individual replace the corresponding genes in CT). If the fitness of a messy-coded individual is higher than the fitness of the CT it is assigned to, then CT is modified (the genes specified by a messy-coded individual replace the appropriate genes in the CT genotype). This paper considers MuPPetS for Flow Assignment in Non-bifurcated Commodity Flow (MuPPetS-FuN) Single that maintains a single CT \cite{mupActive}.\par

    %Although MuPPetS employs a relatively old linkage learning technique that may be imprecise, it seems to improve its effectiveness significantly \cite{muppets}. Such observation was a premise to adjust more up-to-date techniques to WP\_LFL. In \cite{nonBinaryELL}, 3LO dedicated to discrete non-binary search spaces (3LO-nb) was proposed. 3LO-nb can decompose WP\_LFL and reports only the true dependencies. However, same as 3LO, 3LO-nb does not tell direct and indirect dependencies. Nevertheless, the 3LO Algorithm for Flow Assignment in Non-bifurcated Commodity Flow (3LOa-FuN) that uses 3LO-nb reports outstanding results compared to other problem-dedicated optimizers \cite{nonBinaryELL}.

\subsection{Step-like nature}
\label{sec:wpLflf:sln}
One of the main features of WP\_LFL is its \textbf{\textit{step-like nature}}, i.e., to find an optimal solution effectively, we must apply subsequent changes to the best solution found so far, and most of these improving modifications do not improve an initial solution, which should be the easiest to improve.

This statement, that WP\_LFL is of step-like nature, was verified by experiments. We employed MuPPetS-FuN-Single \cite{mupActive}. We considered the same 180 test cases as in \cite{nonBinaryELL}. Six network topologies are used (denoted as 104, 114, 128, 144, 162, and the grid-like network). All networks are built from 36 nodes and can be considered large \cite{netsLarge}. Their minimum and maximum node degrees are between 2 and 6. The experiments can be grouped based on the network they refer to and the experiment type A, B or C. In Group A, we consider 1260 demands (one per pair of nodes), the same capacities of the network arcs and the same size of all demands. In Group B, arc capacities are still equal, but the number of demands is 2500 and connections and demand sizes are chosen randomly. Finally, Group C has the same features as Group B, but the arc capacities may differ. Stop condition based on fitness function evaluation number (FFE) may be unreliable for WP\_LFL \cite{mupActive}. Therefore, each experiment was single-threaded and given three hours of computation time on PowerEdge R430 Dell Server Intel Xeon E5-2670 2.3 GHz 64GB RAM. No other resource-consuming processes were executed, and the number of separate computation processes was one less than the number of physical processor cores.

\begin{table*}[]
 \caption{The number and percentage of improving modifications brought by iterations of MuPPetS-FuN Single that are \textbf{not} applicable to an initial solution}
	\label{tab:modificationNotApplyingToInit}
 %\scriptsize
 \small
\begin{tabular}{l|rrr|rrr||l|rrr|rrr}
& \multicolumn{3}{c}{\textbf{Impr. number}} & \multicolumn{3}{c}{\textbf{Impr. percentage}}  & & \multicolumn{3}{c}{\textbf{Impr. number}} & \multicolumn{3}{c}{\textbf{Impr. percentage}} \\
     & \textbf{Min} & \textbf{Max}  & \textbf{Med}  & \textbf{Min}   & \textbf{Max}   & \textbf{Med} & & \textbf{Min} & \textbf{Max}  & \textbf{Med}  & \textbf{Min}   & \textbf{Max}   & \textbf{Med}  \\
     \hline
\textbf{All}  & 49  & 3876 & 848.5 & 55.06 & 92.25 & 85.13 & \textbf{162}  & 49  & 3713 & 570   & 55.06 & 92.25 & 83.62 \\
\textbf{104}  & 222 & 3519 & 913.5 & 78.45 & 90.58 & 85.48 & \textbf{Grid} & 415 & 3876 & 2280  & 80.43 & 91.22 & 87.00 \\
\textbf{114}  & 76  & 3090 & 705   & 72.68 & 88.69 & 82.65 & \textbf{A}    & 49  & 1027 & 360.5 & 55.06 & 90.44 & 82.39 \\
\textbf{128}  & 332 & 3609 & 1405  & 81.47 & 91.24 & 87.19 & \textbf{B}    & 97  & 3026 & 786   & 74.07 & 91.24 & 87.01 \\
\textbf{144}  & 76  & 2615 & 520.5 & 70.37 & 88.57 & 82.32 & \textbf{C}    & 567 & 3876 & 2397  & 72.97 & 92.25 & 85.19
\end{tabular}
\end{table*}

In our experiments, we were optimizing WP\_LFL instances using MuPPetS-FuN-Single (see Section \ref{sec:wpLflf:opt}). MuPPetS-FuN-Single introduced subsequent modifications to CT (a complete solution to WP\_LFL). We stored each successful improvement. In Table \ref{tab:modificationNotApplyingToInit}, we report the number and percentage of improvements found by MuPPetS-FuN-Single that did \textbf{not} apply to an initial solution. This number is always higher than 50\%, with a median over 80\% and a maximum value over 90\% for most of the considered experiment groups. Thus, most improvements do not improve the initial solution, which should be the easiest to improve.

\section{Leading Ones Problem and its versions}
\label{sec:RelatedWork:leadingOnes}

    One of existing benchmarks being of step-like nature is the \textit{Leading Ones Problem}, which is defined as
    \begin{equation}
        \small
        \mathit{f_{LO}}(\vec{x})=\sum_{i=0}^{n-1} \prod_{j=0}^{i} x_j,
    \end{equation}
    where $\vec{x}=[x_0,x_1,\ldots,x_{n-1}]$ is a binary vector of size $n$. $\mathit{f_{LO}}(\vec{x})$ is frequently used as a benchmark in research considering Single- \cite{SoLeadingOnes} and Multi-Objective optimization \cite{MoGomeaSwarm,moP3} (in the latter case, it is usually paired with the \textit{Trailing Zeroes} problem as the other objective). To solve the Leading Ones Problem, we are to increase the number of subsequent ones at the beginning of the genotype. Such a task may be challenging because the genes after the first zero (counting from the beginning of the genotype) may be considered \textit{disabled}, i.e., do not affect fitness. Oppositely, all genes before the first zero and the gene with the first zero value are \textit{enabled}, i.e., their value affects fitness. Only modifying the first zero value leads to a fitness increase. Additionally, modifying any of the subsequent ones at the beginning of the genotype decreases fitness. Thus, we can state that Leading Ones is of step-like nature, because to find an optimal solution, we must optimize subsequent variables (one after the other) in the appropriate optimization order. \par

    Leading Ones Problem is a base of various benchmark problems. In \cite{leadingOnesDoer}, the Block Leading Ones Problem is proposed and defined as
    \begin{equation}
        \small
        f_{BLO}(\vec{x}) = \lfloor f_{LO}(\vec{x})/l \rfloor.
    \end{equation}
    Such problem can be considered as $\mathit{f_{LO}}(\vec{x})$ with plateaus, i.e., single genes from the Leading Ones Problem are replaced by the blocks of genes (containing $l$ bits) that contribute to fitness only if all bits in the block are set to one. Note that this way of problem construction is well-known \cite{royalRoad}. Another example of the benchmark based on the Leading Ones idea is the Leading Ones Blocks Problem \cite{Lob1,Lob2} defined as
    \begin{equation}
        \small
        \mathit{f_{LOB}(x)}=\sum_{i=0}^{n/b-1} \prod_{j=0}^{b\cdot i - 1} x_j,
    \end{equation}
    which is identical to the Royal Staircase function introduced in \cite{royalStaircase}.

\section{Leading Blocks Problem}
\label{sec:leadingBlocks}

During the experiments described in Section~\ref{sec:wpLflf:sln}, we noticed that the improvements, while optimizing WP\_LFL instances, involve changes made by the sets of many genes instead of single genes as in the Leading Ones. Therefore, this section proposes the Leading Blocks Problem (LBP) for binary search spaces that resemble these step-like optimization features of WP\_LFL. First, we present the motivations and the formal problem definition. Then, we define three types of LBP and discuss their features.\par

\subsection{Motivations and Definition}
Inspired by the features observed during the optimization of the WP\_LFL problem, we propose a benchmark problem that meets the following conditions:\\
\textbf{C1.} Preserves the step-like nature of Leading Ones, i.e., enabling the optimization of various genotype parts after the earlier parts were optimized.\\
\textbf{C2.} Switches from the optimization of single genes to the optimization of gene blocks, where each block is a separate subproblem that may be hard to solve.\\
\textbf{C3.} The disabled blocks (see Section \ref{sec:RelatedWork:leadingOnes}) of genes may also contribute to fitness and create their own variable dependencies that may differ from the dependencies arising from the enabled blocks.\\
\textbf{C4.} The proposed benchmark is more general than the previous propositions inspired by the Leading Ones idea.\\
To meet the above requirements, we define LBP:
    \begin{equation}
    \small
        \label{eq:leadBlocks}
        \mathit{f_{LB}(\vec{x})}=\sum_{s=0}^{S-1} e(s,\vec{x},R) f_s(\vec{x}_{I_s}) + f_d(\vec{x}),
    \end{equation}  
where $I_s$ are disjoint subsets of $\{0,...,n-1\}$, $S$ is the number of subsets, $f_s$ is a subfunction, $e(s,\vec{x},R)$ is a function returning 1 if the $s$-th subfunction is enabled or 0 otherwise, and $f_d$ is a function that defines the influence of the genes in the disabled blocks on the fitness value (in this subsection, we consider $f_d(\vec{x}) = 0$; other cases are considered in Section \ref{sec:ledingBlocks:types}). Function $e(s,\vec{x},R)$ is defined as follows.
    \begin{equation}
        \label{eq:leadEnabled}
    \small
        \mathit{e(s,\vec{x},R)}=
        \begin{cases}
            1,s < R \text{  }\lor \text{  }\bigl( \exists_{i \in \{s-R,...,s-1\}}e(i,\vec{x},R)=1 \text{  }\land f_i(\vec{x}_{I_i}) = f_{i,opt} \bigr), \\
            0, \text{otherwise}
        \end{cases}
    \end{equation}
where $R$ is a user-defined parameter, and $f_{i,opt}$ is the optimal value of the i\textit{th} block.\par

Note that $e(s,\vec{x},R)$ always marks the first $R$ blocks as enabled. The other blocks are enabled if at least one of the $R$ preceding blocks is enabled and its fitness value is optimal. Finally, the value of $f_{LB}(\vec{x})$ is the sum of $f_s$ values of all enabled blocks, and $f_d$ computed for all genes that belong to disabled blocks.\par

The LBP definition is more general than the definitions of $f_{LO}$, $f_{BLO}$, and $f_{LOB}$ presented in Section \ref{sec:RelatedWork:leadingOnes}. For instance, if we consider LBP with $R=1$, 
%\sout{$f_s(\vec{x})=u(\vec{x}_{I_s})$, $f_d(\vec{x}) = 0$, and $I_s=\{s\}$, where $u(\vec{x})$ is the unitation of $\vec{x}$ (the sum of binary values \cite{3lo})} 
$I_s=\{s\}$, $f_s(\vec{x}_{I_s})=x_s$ and $f_d(\vec{x}) = 0$, we obtain an instance of Leading Ones. Similar examples can be defined of $f_{BLO}$ and $f_{LOB}$.\par

Let us now consider the LBP instance where $n=16$, $R=1$, $I_s=\{4 \cdot s, 4 \cdot s + 1, 4 \cdot s + 2, 4 \cdot s + 3\}$, $f_d(\vec{x}) = 0$, and $f_s(\vec{x}_{I_s})=bimTrap_4(u(\vec{x}_{I_s}))$. $bimTrap_4$ is a bimodal deceptive function of unitation of order $k = 4$ \cite{3lo,decBimodalOld} defined as follows.
%\begin{equation}
%\label{eq:decTrap}
%\mathit{decTrap}_k(u) = 
%\begin{cases}
%k - 1 - u & ,u \neq k \\
%k  & ,u = k \\
%\end{cases}
%\end{equation}
\begin{equation}
\small
\label{eq:bimodal}
\mathit{bimTrap}_k(u) = 
\begin{cases}
k / 2 - |u - k/2| - 1 & ,u \neq k \land u \neq 0\\
k / 2 & ,u = k \lor u = 0\\
\end{cases},
\end{equation}
where $u$ is the unitation (the sum of binary values) of $\vec{x}$.\\
For the above problem, we consider $\vec{x}_a=[1111\textbf{ }0110\textbf{ }0000\textbf{ }0010]$. The first and the third blocks are defined on genes 0-3 and 8-11, respectively. These blocks are optimal (the value of a bimodal function is optimal if the argument contains only 1s or only 0s). The first block is enabled because $s=0 < R=1$. The second block is enabled because the first block is enabled, and $f_0(\vec{x}_{a,I_0}) = f_{0,opt} = 2$. The third block is disabled because the value of the (enabled) second block is suboptimal. Finally, the fourth block is disabled because the third block is disabled. Thus, $f_{LB}(\vec{x}_a)=f_0(u(1111)) + f_1(u(0110)) + f_d(\vec{x}_a) = 2 + 1 + 0 = 3$.\par

In the above example, if we consider $R=2$, then all blocks are enabled. The first and second blocks are enabled because $s<R$ for $s=0$ and $s=1$, respectively. The third block is enabled because one of the two preceding blocks is optimal (the first block), and the fourth block is enabled because one of the blocks 2 and 3 is enabled and optimal (the third block). Therefore, in such a case, $f_{LB}(\vec{x}_a)=f_0(u(1111)) + f_1(u(0110)) + f_2(0000) + f_3(0010) = 2 + 1 + 2 + 0 = 5$.
\par

\subsection{The Basic Features of Leading Blocks Problem}
	\label{sec:ledingBlocks:features}

We can interpret the subsequent states of an optimization process as the sequence of improvements (modifications) of the best-found solution. Each modification may be applied earlier on the way, but it may turn out that it does not increase fitness then. However, from the history of modifications, one can take out such a sequence that each step improves fitness. The length of the shortest such sequence of improvements leading to the situation that the current modification improves the fitness will be denoted as \textit{\textbf{the number of hops}} from the initial state to the current modification. The number of hops can be defined as follows.\par

 Let $X$ be the set of $b$ best-found individuals at the subsequent optimization stages, and let $\mathcal{T}=(T_1,...,T_b)$ be a sequence of self-maps $T_i:X\to X$, $i=1,...,b$. The maps $T_i$ are interpreted as the consecutive modifications of the best-found individual made during the optimization process. This process is represented by a sequence $\vec{x}_0,\vec{x}_1,...,\vec{x}_b$ of individuals satisfying $f(\vec{x}_{i-1})<f(\vec{x}_i)$, where $\vec{x}_0$ is the initial point and $\vec{x}_i=T_i\circ T_{i-1}\circ...\circ T_1(\vec{x}_0) = T_i(T_{i-1}(...T_1(\vec{x}_0)))$.
For convenience, by $T_0$ we denote the identity map (interpreted as ``no change'').
By the \textit{number of hops} leading to $T_b$ in $\vec{x}_0$ through $\mathcal{T}$ we mean the smallest number $h$ such that there is a subsequence $T_{i_1},...,T_{i_h}$ with $i_h=b$ and 
\begin{equation}
\small
\label{eq:hops}
\begin{aligned}
\forall&\ j=1,...,h \\
&\qquad f\big(T_{i_j}(T_{i_{j-1}}(...T_{i_1}(T_{i_0}(\vec{x}_0))))\big) > f\big(T_{i_{j-1}}(...T_{i_1}(T_{i_0}(\vec{x}_0)))\big).
\end{aligned}
\end{equation}
\begin{algorithm}
	\caption{The Algorithm for the Hop Number Estimation}
    %\scriptsize
	\begin{algorithmic}[1]
        \Function{EstimateHOPs}{$bInds$}
            \State $hops \gets 0$;
            \State $cur \gets $ sizeof($bInds$) - 1; \label{line:hops:lastInd}
            \While {$cur > 0$}
                \State $last \gets cur-1$
                \State $mod \gets$ GetMod($bInds[cur]$, $bInds[last]$); \label{line:hops:mod}
                \While {Fit(Mod($bInds[last-1]$, $mod$))$ > $ Fit($bInds[last-1]$) \textbf{AND} $last > 0$} 
    			 \State $last \gets last - 1$;
                \EndWhile
                \State $hops \gets hops+1$;
			\EndWhile
            \Return{$hops$}
		\EndFunction
	\end{algorithmic}
	\label{alg:hops}
\end{algorithm}

The exact computation of the hop number may be computationally expensive. Therefore, to estimate the number of hops, we use the algorithm that finds
its upper bound (Pseudocode \ref{alg:hops}). $bInds$ is a set of individuals containing subsequent best-found individuals obtained during optimization. The index of an initial individual is 0. We start from the last best-found individual (line \ref{line:hops:lastInd}), which is a result of the optimization process. We obtain the modification for the current individual (line \ref{line:hops:mod}). For instance, if for the 6-bit problem $bInds[cur] = 111000$ and $bInds[last] = 110100$, then the modification $mod = **01**$. Then, we check for how many individuals that precede $bInds[cur]$ fitness increases after applying the modification. The modification is applied to an individual in the following way. For the modification $mod = **01**$ and the individual $ind=000000$, the modified individual is Mod($ind$)=$000100$. When we can not apply the modification any further, then we increase the number of hops by one.\par

 \begin{table*}[]
 \caption{The optimization steps of an exemplary solution to the LBP instance built from four $\bm{bimTrap_4}$ functions (modified blocks are marked in bold and italic). The last column presents the hops for the problem built from the concatenation of four $\bm{bimTrap_4}$ functions.}
	\label{tab:optimizationExample:leading}
    % \scriptsize
\small
\begin{tabular}{rrrrr|rrrr|r|l|r|l}
     \multicolumn{5}{c}{\textbf{Solution}} & \multicolumn{4}{c}{\textbf{Modification}}  & $\bm{f_{LB}}$ & \textbf{Hops $\bm{f_{LB}}$} & $\bm{f_{con}}$ & \textbf{Hops $\bm{f_{con}}$}  \\
     \hline
         $\vec{x}_0$ & 1011 & 1010 & 0100 & 0111  &  \multicolumn{4}{c}{N/A}  & 0 & N/A & 1 & N/A\\
         $\vec{x}_1$ & \textit{\textbf{0101}} & 1010 & 0100 & 0111    &    010* & **** & **** & **** & 1 & 1 (1$\rightarrow$0) & 2 & 1 (1$\rightarrow$0) \\
         $\vec{x}_2$ & \textit{\textbf{1111}} & 1010 & 0100 & 0111    &    1*1* & **** & **** & **** & 3 & 2 (2$\rightarrow$1$\rightarrow$0) & 3 & 2 (2$\rightarrow$1$\rightarrow$0)\\
         $\vec{x}_3$ & 1111 & \textit{\textbf{1111}} & 0100 & 0111    &    **** & *1*1 & **** & **** & 4 & 3 (3$\rightarrow$2$\rightarrow$1$\rightarrow$0) & 4 & 1 (3$\rightarrow$0) \\
         $\vec{x}_4$ & 1111 & 1111 & \textit{\textbf{0101}} & 0111    &    **** & **** & ***1 & **** & 5 & 4 (4$\rightarrow$3$\rightarrow$2$\rightarrow$1$\rightarrow$0) & 5 & 1 (4$\rightarrow$0) \\
         $\vec{x}_5$ & 1111 & 1111 & \textit{\textbf{1111}} & 0111    &    **** & **** & 1*1* & **** & 6 &  5 (5$\rightarrow$4$\rightarrow$3$\rightarrow$2$\rightarrow$1$\rightarrow$0) & 6 & 2 (5$\rightarrow$4$\rightarrow$0) \\
         $\vec{x}_6$ & 1111 & 1111 & 1111 & \textit{\textbf{1111}}    &    **** & **** & **** & 1*** & 8 & 6 (6$\rightarrow$5$\rightarrow$4$\rightarrow$3$\rightarrow$2$\rightarrow$1$\rightarrow$0) & 8 & 1 (6$\rightarrow$0) \\        
\end{tabular}
\end{table*}

Let us analyze an example of subsequent improvements of a solution to the LBP instance where $n=16$, $R=1$, $I_s=\{4 \cdot s, 4 \cdot s + 1, 4 \cdot s + 2, 4 \cdot s + 3\}$, $f_d(\vec{x}) = 0$, and $f_s(\vec{x}_{I_s})=bimTrap_4(u(\vec{x}_{I_s}))$. In Table \ref{tab:optimizationExample:leading}, we start from solution $\vec{x}_0$ and present the subsequent optimization steps leading to the optimal solution. Each step finishes in a locally optimal solution. Initially, only the first block is enabled and influences fitness. The first optimization step (from $\vec{x}_0$ to $\vec{x}_1$) improves the quality of the first block ($bimTrap_4(u(1011)) = 0$, while $bimTrap_4(u(0101)) = 1$). In the second step, the first block is improved again and becomes optimal. Note that the second step would not improve $\vec{x}_0$. It applies only to $\vec{x}_1$, i.e., it improves the fitness of $\vec{x}_1$ but does not improve the fitness of $\vec{x}_0$. Therefore, column \textit{Hops $f_{LB}$} states that the modification performed in step 2 is two hops away from the initial solution $\vec{x}_0$. The first block in $\vec{x}_2$ is optimal. Therefore, the second block in $\vec{x}_2$ is enabled and influences fitness. Step 3 improves the fitness of the second block, makes it optimal, and enables the third block. Steps 4 and 5 are similar to steps 1 and 2 -- they improve the fourth block by finding its locally optimal value first ($0101$) and then finding the optimal one ($1111$). In both cases, the hop-based distance from the initial solution is increased by two. This is a result of a fitness landscape feature of a single block (it is improved to a locally optimal solution first, then to the globally optimal one). Finally, steps 5 and 6 set the optimal block values to blocks 3 and 4, respectively.\par

The above scenario refers to LBP. However, the same optimization steps would apply to the concatenation of four $bimTrap_4$ functions defined as $\mathit{f_{con}(\vec{x})}=\sum_{s=0}^{S-1} f_s(\vec{x}_{I_s})$.
%Note that the genotype length of $\mathit{f_{conc}(x)}$ problem is the same as $\mathit{f_{LB}(x)}$. Moreover the same optimization steps will lead the optimization of $x_0$ to the optimal solution. 
Therefore, column \textit{Hops $f_{con}$} reports the number of hops from a given solution to $\vec{x}_0$ for the $f_{con}$ problem. As shown, the same optimization steps improve the subsequent solutions for LBP and $f_{con}$. However, the number of hops to $\vec{x}_0$ is significantly lower for the latter problem. 
%It is higher than one only when two improvement steps refer to the same block of genes. The situation is the opposite for LBP -- the number of hops increases when the latter blocks are improved because the preceding blocks must be optimized first. 
This example shows the main difference between the concatenation of subfunctions and LBP. In the case of concatenation, we can optimize any block at any optimization stage, while for LBP, we can optimize only the enabled blocks. \par

\begin{table}[]
 \caption{The optimization steps of an exemplary solution to the Cyclic-trap built from four $\bm{bimTrap_4}$ functions and overlap $\bm{o=1}$ (modified blocks are marked in bold and italics)}
 %(modified blocks are marked in bold and italics)}
	\label{tab:optimizationExample:cyclic}
    % \scriptsize
\small
\begin{tabular}{l|r|rrrr|l}
     \textbf{Step} & \textbf{Solution} & \multicolumn{4}{c}{\textbf{Blocks}}    &  \textbf{Hops}  \\
     \hline
        0 & $\vec{x}_0$ 101101010011  & 1011 & 1010 & 0100 & 0111  & N/A \\
        1 & $\vec{x}_1$ \textit{\textbf{1111}}01010011 & \textit{\textbf{1111}} & 1010 & 0100 & 0111    &   1 (1$\rightarrow$0) \\
        2 & $\vec{x}_2$ 11\textit{\textbf{11111100}}11 & 1111 & \textit{\textbf{1111}} & \textit{\textbf{1100}} & 0111    & 1 (2$\rightarrow$0) \\
        3 & $\vec{x}_3$ \textit{\textbf{1}}11111110\textit{\textbf{001}} & 1111 & 1111 & 1100 & \textit{\textbf{0011}}    & 1 (3$\rightarrow$0) \\
        4 & $\vec{x}_4$ \textit{\textbf{1}}11111\textit{\textbf{111111}} & 1111 & 1111 & \textit{\textbf{1111}} & \textit{\textbf{1111}}     & 1 (4$\rightarrow$0)
\end{tabular}
\end{table}

In Table \ref{tab:optimizationExample:cyclic}, we analyze the optimization steps for the cyclic trap problem using $bimTrap_4$ blocks and overlap $o=1$ \cite{dsmga2}. In cyclic traps built from functions of order $k$, the first block is defined by genes $[1;k]$. The next block is defined by genes $[k-o; 2\cdot k-o]$, etc. Each block shares $o$ of its genes with its predecessor and $o$ of its genes with the next block. The first and the last block also share $o$ of their genes. Similarly to the example with $f_{con}$, the number of hops is low because all considered improvements can be applied to the initial solution $\vec{x_0}$. 

%In Section \ref{sec:expWPlfl}, we show that the considered instances of the WP\_LFL problem can be improved iteratively, but many of these improvements are characterized by a high number of hops (many prior improvements must be made to enable the latter ones). Similarly, we show that the benchmarks that are frequently considered in GA-based literature can be improved iteratively, but the number of hops for almost all of these improvements is low (frequently, it equals one). Therefore, proposing the Leading Blocks Problem fills the gap in the benchmark set that is available in the GA-based literature.

\subsection{Considered Leading Blocks Problem Types}
	\label{sec:ledingBlocks:types}

We propose three types of LBP that differ by the definition of $f_d$. For the first type, \textit{RestOff}, $f_d(\vec{x})=0$. For the second type, \textit{HalfOnHalf}, $f_d$ is defined as

%\begin{equation}
%\small
%    \label{eq:resOffFd}
%    f_d(\vec{x})= 
%    \begin{cases}
%    (f_{dmax}(\vec{x})\cdot 0.1) \cdot \frac{1-|u_d-size_d/2|}{size_d/2} & ,u_d > 0\\
%    0 & ,otherwise\\
%    \end{cases}
%\end{equation}  

\begin{equation}
\small
    \label{eq:resOffFd}
    f_d(\vec{x})= (f_{dmax}(\vec{x})\cdot \alpha) \cdot \Bigg(1-\frac{|u_d-size_d/2|}{size_d/2}\Bigg),
\end{equation}  
where $f_{dmax}(\vec{x})=\sum_{s=0}^{S-1} (1-e(s,\vec{x},R)) \cdot f_{s,opt}$ is the sum of optimal values of all disabled blocks, $\alpha \in (0;1)$ is a constant value, $size_d$ is the number of genes in the disabled blocks, and $u_d$ is the unitation of all of the disabled blocks. 

%In the third type of considered Leading Problems, \textit{Alter}, $f_d(x)$ is defined as
Finally, in the third type, denoted as \textit{Alter}, $f_d(\vec{x})$ is defined as
\begin{equation}
\small
    \label{eq:leadAlterFd}
    \mathit{f_d(\vec{x})}=\sum_{s=0}^{S-1} (1-e(s,\vec{x},R)) g_s(\vec{x}_{I_s}),
\end{equation}  
where $g_s$ is an alternative function to $f_s$. In this paper, as $f_s$, we use $bimTrap_{10}$. Therefore, we use $g_s$ as a bimodal function without its optima:
\begin{equation}
\small
    \label{eq:leadAlterBimodal}
        g_s(u(\vec{x}_{I_s})) = noOptBimodal_k(u(\vec{x}_{I_s})) = k / 2 - |u(\vec{x}_{I_s}) - k/2| - 1. \\
\end{equation} 
%\begin{equation}
%\small
 %   \label{eq:leadAlterBimodal2}
 %       noOptBimodal_k(u) = k / 2 - |u - k/2| - 1
%\end{equation} 

    All three LBP types have the same dependency structure for the enabled blocks. However, the dependencies for genes that are a part of disabled subfunctions differ. For clarity, we will denote the genes as \textit{disabled genes} if they are a part of disabled subfunctions and \textit{enabled genes} in the other case.\par
    
    %According to the DLED-like definition of dependency, for the \textit{RestOff} type of the Leading Blocks Problem, the disabled genes are independent. For the \textit{HalfOnHalf} type, they are all directly dependent on each other. Finally, for the \textit{Alter} type, the dependencies between the disabled genes are the same as for the enabled genes. These differences cause differences in the effectiveness of both considered LT-GOMEA versions.\par

\section{Hop-based Analysis of WP\_LFL}
\label{sec:expWPlfl}

\begin{figure}[]
    \begin{subfigure}[b]{1.0\linewidth}
        \begin{center}

		\resizebox{0.9\linewidth}{!}{%
			\centering
	       \includegraphics[width=0.95 \linewidth]{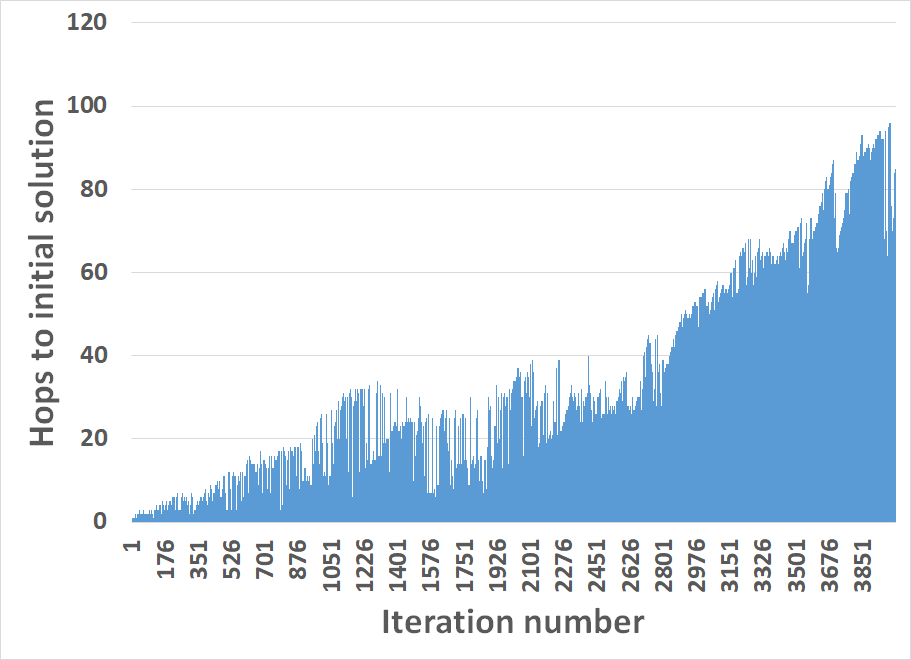}
	
		}
              \end{center}
		\caption{HOPs to an initial solution based on modifications brought by subsequent iterations of MuPPetS-FuN Single; network 162, experiment group C}
	   \label{fig:hops:hopsToInitial}

	\end{subfigure}
    \begin{subfigure}[b]{1.0\linewidth}
        \begin{center}
            
		\resizebox{0.7\linewidth}{!}{%
			\tikzset{every mark/.append style={scale=2.5}}
			\begin{tikzpicture}
			\begin{axis}[%
            legend entries={\textbf{A}, \textbf{B}, \textbf{C}, \textbf{All}},
            legend pos=north west,
			xtick={1,2,3,4,5,6,7,8,9,10,11,12,13,14,15,16,17,18,19,20},
			xticklabels={1,,,,5,,,8,,,11,,,14,,,17,,,20+},
			xmin=0.5,
			xmax=20.5,
			%ymode=log,
			%ymin=1e4,
			%ymax=1e9,
			%legend pos=south east,
			xlabel=\textbf{Hops},
			ylabel=\textbf{Modifications},
			grid,
			grid style=dashed,
			ticklabel style={scale=0.8},
			label style={scale=1.0},
			legend style={font=\fontsize{8}{0}\selectfont}
			]

            \addplot[
			color=red,
			]
			coordinates {
				(1,47.5)(2,41.5)(3,35.5)(4,32.5)(5,25)(6,22.5)(7,20)(8,15.5)(9,13)(10,10.5)(11,10)(12,9)(13,6)(14,5)(15,4)(16,3.5)(17,3.5)(18,3)(19,3)(20,12)
			};

            \addplot[
			color=darkgreen,
			]
			coordinates {
				(1,83.5)(2,74.5)(3,68.5)(4,62)(5,53)(6,46.5)(7,38.5)(8,35)(9,35)(10,31.5)(11,28.5)(12,22.5)(13,19)(14,21)(15,18.5)(16,17)(17,12.5)(18,12)(19,9.5)(20,104.5)

			};

            \addplot[
			color=blue,
			]
			coordinates {
				(1,270.5)(2,197.5)(3,171)(4,159)(5,135.5)(6,124.5)(7,104)(8,89)(9,80.5)(10,66.5)(11,70.5)(12,55.5)(13,49.5)(14,47)(15,44.5)(16,35)(17,31.5)(18,27.5)(19,32.5)(20,435)
			};

            \addplot[
			color=black,
			]
			coordinates {
				(1,84.5)(2,75)(3,73)(4,64.5)(5,56)(6,48.5)(7,42.5)(8,42)(9,38)(10,30.5)(11,27.5)(12,22)(13,19)(14,19)(15,16.5)(16,15)(17,12)(18,10)(19,10)(20,113)

			};

			\end{axis}
			\end{tikzpicture}
		}
        \end{center}
		\caption{Median number of hops to an initial solution per modification for various test case groups}
	   \label{fig:hops:hopsGeneral}
	\end{subfigure}
	
	\caption{HOP-based analysis of the WP\_LFL instances}
	\label{fig:hops}
\end{figure}

This section presents the HOP-based analysis (see Section \ref{sec:ledingBlocks:features}) of WP\_LFL instances. The setup of the experiments was the same as described in Section~\ref{sec:wpLflf:sln}. Moreover, we also calculated the number of hops of each successful improvement of CT (a complete solution to WP\_LFL). In Figure \ref{fig:hops:hopsToInitial}, we show the results of a representative experiment showing the number of hops for the subsequent improvements of the best-found solution found by MuPPetS-FuN Single. The number of hops for most improvements is high and increases with the increase of the iteration number. This shows that the improvements in the latter iterations require many improvements that were made before. Thus, one improvement enables the next improvement and so on. Figure \ref{fig:hops:hopsGeneral} shows the median number of improvements that are a given number of hops away from an initial solution. Note that most improvements for experiment groups B and C are more than 20 hops away from the initial solution. \par

\section{Typical Benchmarks}

In this section, we first introduce state-of-the-art optimizers that can solve typical binary benchmarks effectively. Some of these optimizers are used during our experiments to perform the HOP-based analysis of typical benchmarks. The analysis is presented at the end of this section.

 \subsection{Problem decomposition in state-of-the-art GA-based optimizers}
\label{sec:RelatedWork:llGAs}

Using the knowledge about variable dependencies is a way to improve GAs' effectiveness. For instance, in \cite{whitley2019}, the authors state that such information should be exploited whenever possible. It is frequent to cluster the most dependent variables and use these clusters in the mixing (crossover-like) operations \cite{dsmga2,P3Original,3lo,ltga,pxForBinary}. Note that the clusters of dependent variables can be utilized in other ways than individual mixing. For instance, we can use such clusters as perturbation masks in the Iterated Local Search (ILS) \cite{ilsDLED}.\par

Statistical Linkage Learning (SLL) aims to discover dependencies by statistically analyzing the frequencies of gene value pairs in the GA population \cite{ltga,P3Original,dsmga2}. Mutual information is frequently used \cite{mutualInformation}: $ I(X, Y) = \sum_{x \in X} \sum_{y \in Y} p(x, y) \log_2 \frac{p(x,y)}{p(x)p(y)}$, where $X$ and $Y$ are two random variables (genes). The entropy values (referring to all pairs of genes) are stored in the Dependency Structure Matrix (DSM). Linkage Trees (LT) are frequently used to cluster DSM. Such clusters can be used by mixing operators \cite{ltga,P3Original,3lo}. LT is constructed starting from its leaves (each leaf is a cluster referring to a single gene). Then, the most dependent clusters (according to DSM) are joined until we obtain a cluster containing all genes (the root of LT).\par

% \begin{equation}
%     I(X, Y) = \sum_{x \in X} \sum_{y \in Y} p(x, y) \log_2 \frac{p(x,y)}{p(x)p(y)}
% \end{equation}

    Optimal Mixing (OM) uses LT and is a part of some of the state-of-the-art GAs \cite{P3Original,ltga,dled,3lo,FIHCwLL,dgga}. OM considers two individuals (\textit{source} and \textit{donor}) and a mask. Genes marked by a mask are copied from the donor to the source individual. If the fitness of the source individual has not decreased, the modification is preserved or reverted otherwise. OM is a part of the LT Gene-pool Optimal Mixing Evolutionary Algorithm (LT-GOMEA) \cite{permutLTGOMEA,dled}. LT-GOMEA is similar to standard GA but uses OM instead of crossover, mutation, and selection. It employs the population-sizing scheme \cite{HarikPopSizing} that removes the necessity of specifying the population size and makes LT-GOMEA parameter-less.\par

    SLL-using optimizers were shown highly effective in solving many real-world and benchmark problems \cite{P3Original,permutLTGOMEA}. However, SLL-based decomposition may be imprecise or even support false information when it tries to decompose some problems \cite{3lo}. Additionally, some studies show that the low quality of problem decomposition may cause SLL-using optimizers to be ineffective \cite{linkageQuality}. Therefore, Empirical Linkage Learning (ELL) techniques were proposed. ELL is proven to discover only the true dependencies, although it does not guarantee their discovery. Direct Linkage Empirical Discovery (DLED) is one of the recent ELL propositions that is proven to discover only the direct variable dependencies. Thus, for the problems of additive nature \cite{irrg}, DLED discovers only those variable dependencies that will be obtained by the Walsh decomposition \cite{grayBoxMkLand,grayBoxGecco,dgga}. This feature enabled the use of the Gray-box mechanisms in Black-box optimization \cite{dgga} (in Gray-box optimization, the complete set of direct variable dependencies is supported by a user and utilized in different forms during the optimization). For the binary search space, DLED discovers that two variables are dependent if the following condition holds: $(f(\vec{x}) < f(\vec{x}^g) \land f(\vec{x}^h) \geq f(\vec{x}^{g,h}) )\lor  (f(\vec{x}) \geq f(\vec{x}^g) \land   f(\vec{x}^h) \geq f(\vec{x}^{g,h}))$, where $\vec{x}^{g}$, $\vec{x}^h$, $\vec{x}^{g,h}$, denote the individuals obtained from $\vec{x}$ by flipping gene $g$, gene $h$ or both, respectively.

    % \begin{equation}
    %     \begin{cases}
    %         f(\vec{x}) < f(\vec{x}^g) \\
    %         f(\vec{x}^h) \geq f(\vec{x}^{g,h})
    %     \end{cases} \lor \begin{cases}
    %         f(\vec{x}) \geq f(\vec{x}^g) \\
    %         f(\vec{x}^h) \geq f(\vec{x}^{g,h})
    %     \end{cases}
    % \end{equation}

    %\begin{equation}
    %\small
    %\label{link_check}
     %  \Big(f(\vec{x}) < f(\vec{x}^g) \text{\textbf{ AND }} f(\vec{x}^h) \geq f(\vec{x}^{g,h})\Big) \text{\textbf{ OR }} \Big(f(\vec{x}) \geq f(\vec{x}^g) \text{\textbf{ AND }} f(\vec{x}^h) < f(\vec{x}^{g,h})\Big)
    %\end{equation}

    Similarly to monotonicity checking strategies dedicated to continuous search spaces, DLED can decompose additive and non-additive problems, which is a significant advantage \cite{irrg}. Its computational cost is low enough to apply it to LT-GOMEA \cite{dled}. LT-GOMEA-DLED was shown to be more effective than the original SLL-using LT-GOMEA for those problems for which SLL does not support a decomposition of sufficiently high quality.

\subsection{Hop-based Analysis}
\label{sec:BenchhopBasedAnal}

In Section~\ref{sec:expWPlfl}, we present the results considering the hop-like nature of the considered WP\_LFL test cases. To the best of our knowledge, the hop-like nature of optimization problems frequently considered in the research on evolutionary computation has not been investigated yet. Therefore, we propose the following experiment. We propose the Iterated Local Search with SLL (ILS-SLL). In each iteration, ILS-SLL randomly creates a solution and optimizes it with the First Improvement Hillclimber (FIHC), which is a local search algorithm \cite{P3Original,dsmga2,3lo,FIHCwLL}. Then, it creates a DSM and builds LT in the same way as LT-GOMEA. For this purpose, it uses the set of solutions created in the previous iterations. LT clusters serve as perturbation masks. After perturbation, ILS-SLL optimizes the solution with FIHC. If fitness increases, the changes are preserved or rejected otherwise. All improving modifications are stored, and we analyze their hop-like nature, i.e., how many hops from an initial solution a given improvement is.\par

We consider the well-known optimization problems for the binary domains -- deceptive functions concatenations (including standard deceptive, bimodal deceptive and noised bimodal deceptive functions) \cite{3lo,dled}, Mk Landscape problem using $bimTrap_{10}$ \cite{grayBoxMkLand,mkLandDirk}, Max3Sat \cite{grayBoxMkLand,P3Original,dgga}, and Ising Spin Glasses (ISG) \cite{P3Original,dgga}. The experiment setup was the same as in the previous section but with a computation time of 30 minutes. Each experiment was repeated 30 times. Since these problems are well-known, we only refer to papers containing their definitions.\par

\begin{table*}[]
 \caption{The hop-like nature of the frequently considered optimization problems on the base of ILS-SLL runs}
	\label{tab:bench:hopNature}
 % \scriptsize
 \small
\begin{tabular}{l|rrrrrrr||l|rrrrrrr}
                    &      &      & \multicolumn{2}{c}{\textbf{Hops}} & \multicolumn{3}{c}{\textbf{$\mathbf{Hops_{max}}$}} &      &      & \multicolumn{2}{c}{\textbf{Hops}} & \multicolumn{3}{c}{\textbf{$\mathbf{Hops_{max}}$}}      \\
                    &  \textbf{\textit{n}}    & \textbf{Mods} & \textbf{Avr}  & \textbf{StD} & \textbf{Med}    & \textbf{Min} & \textbf{Max} & &  \textbf{\textit{n}}    & \textbf{Mods} & \textbf{Avr}  & \textbf{StD} & \textbf{Med}    & \textbf{Min} & \textbf{Max}\\
                    \hline
\textbf{{bim10}}         & 400  & 65   & 1.21 & 0.44  & 2.0       & 2   & 3   & \textbf{mk-}  & 143  & 23   & 1.17 & 0.48  & 2.0       & 1   & 3   \\
                    & 800  & 90   & 1.14 & 0.37  & 2.0       & 2   & 3   & \textbf{bim10}                   & 283  & 46   & 1.20 & 0.46  & 2.0       & 2   & 3   \\
                    & 1200 & 134  & 1.12 & 0.35  & 2.0       & 2   & 3   & \textbf{o=3} & 563  & 83   & 1.21 & 0.45  & 2.0       & 2   & 3   \\
                    & 1600 & 178  & 1.14 & 0.38  & 2.5     & 2   & 3   & \textbf{b=2} & 843  & 122  & 1.25 & 0.47  & 3.0       & 2   & 3   \\
\textbf{{bimN10}}   & 400  & 26   & 1.10 & 0.40  & 2.0       & 1   & 3   & \textbf{Max-}             & 100  & 34   & 1.46 & 0.64  & 3.0      & 2   & 4   \\
                    & 800  & 46   & 1.14 & 0.41  & 2.0       & 2   & 3   & \textbf{3Sat} & 200  & 64   & 1.45 & 0.59  & 3.0       & 2   & 4   \\
                    & 1200 & 65   & 1.21 & 0.46  & 2.0       & 2   & 3   & & 500  & 161  & 1.50 & 0.59  & 3.0       & 3   & 4   \\
                    & 1600 & 83   & 1.19 & 0.44  & 2.0       & 2   & 4   & & 1000 & 315  & 1.49 & 0.57  & 3.0       & 3   & 4   \\
\textbf{{dec8}}              & 400  & 213  & 1.10 & 0.31  & 2.0       & 2   & 3   & \textbf{ISG}                 & 484  & 131  & 1.36 & 0.53  & 3.0       & 2   & 4   \\
                    & 800  & 395  & 1.13 & 0.35  & 2.5     & 2   & 3   & & 625  & 169  & 1.39 & 0.54  & 3.0       & 2   & 4   \\
                    & 1200 & 540  & 1.17 & 0.38  & 3.0       & 2   & 3   & & 784  & 208  & 1.40 & 0.53  & 3.0       & 3   & 4   \\
                    & 1600 & 712  & 1.17 & 0.38  & 3.0       & 2   & 3   & & 1521 & 396  & 1.37 & 0.53  & 3.0       & 3   & 4 
					
\end{tabular}
\end{table*}

\begin{figure*}[]
    \begin{subfigure}[b]{0.32\linewidth}
		\resizebox{0.8\linewidth}{!}{%
			\tikzset{every mark/.append style={scale=2.5}}
			\begin{tikzpicture}
			\begin{axis}[%
            legend entries={LT-GOMEA-DirLink, LT-GOMEA-DLED, LT-GOMEA},
            legend columns=-1,
            legend to name=named,
			%legend columns=-1,
            %legend entries={dgGA,cGOMEA,P3,P3-DLED,LT-GOMEA-DLED},
            %legend to name=named,
            %legend columns=-1,
            %legend entries={LT-GOMEA-DLED,LT-GOMEA-SLL,P3-DLED,P3-SLL},
            %legend to name=named,
			xtick={1,2,3,4,5,6},
			xticklabels={50,70,100,150,200,300},
			xmin=0.5,
			xmax=6.5,
			ymode=log,
			%ymin=1e4,
			%ymax=1e9,
			%legend pos=south east,
			%xlabel=\textbf{unitation},
			ylabel=\textbf{Fitness},
			grid,
			grid style=dashed,
			ticklabel style={scale=1.5},
			label style={scale=1.5},
			legend style={font=\fontsize{8}{0}\selectfont}
			]

            %\addplot[
			%color=brown,
			%mark=triangle*,
			%]
			%coordinates {
			%	(1,477026)(2,1961269)(3,7857276.5)(4,11729329)(5,15856165)(6,30264820)(7,24703945)(8,32664404)
			%};
   
            %\addplot[
			%color=black,
			%mark=*,
			%]
			%coordinates {
            %    (1,439884.5)(2,1464475.5)(3,4787594)(4,7602447)(5,13019899.5)(6,22470466)(7,27529834)(8,37681398)
			%};

            \addplot[
			color=black,
			mark=triangle*,
			]
			coordinates {
				(1,3003236.5)(2,7125243)(3,27033777.5)(4,45996170.5)(5,114912849.5)(6,274582111)
			};
   
            \addplot[
			color=blue,
			mark=*,
			]
			coordinates {
				(1,2779283.5)(2,6759286)(3,15241144)(4,37471604)(5,73148461)(6,255850222)
			};

            \addplot[
			color=red,
			mark=triangle*,
			]
			coordinates {
				(1,587173.5)(2,1236369.5)(3,2388064.5)(4,3903832.5)(5,9563556)(6,15939334.5)
			};
   
			\addplot[
			color=violet,
			mark=*,
			]
			coordinates {
				(1,947494.5)(2,1590117.5)(3,2401516)(4,5904031)(5,8615570)(6,15855735)
			};

			\end{axis}
			\end{tikzpicture}
		}
		\caption{\textit{RestOff} $\bm{R=1}$}
		\label{fig:leading:floating:1}
	\end{subfigure}
    \begin{subfigure}[b]{0.32\linewidth}
		\resizebox{0.8\linewidth}{!}{%
			\tikzset{every mark/.append style={scale=2.5}}
			\begin{tikzpicture}
			\begin{axis}[%
            legend entries={LT-GOMEA-DirLink, LT-GOMEA-DLED, LT-GOMEA},
            legend columns=-1,
            legend to name=named,
			%legend columns=-1,
            %legend entries={dgGA,cGOMEA,P3,P3-DLED,LT-GOMEA-DLED},
            %legend to name=named,
            %legend columns=-1,
            %legend entries={LT-GOMEA-DLED,LT-GOMEA-SLL,P3-DLED,P3-SLL},
            %legend to name=named,
			xtick={1,2,3,4,5,6},
			xticklabels={50,70,100,150,200,300},
			xmin=0.5,
			xmax=6.5,
			ymode=log,
			%ymin=1e4,
			%ymax=1e9,
			%legend pos=south east,
			%xlabel=\textbf{unitation},
			ylabel=\textbf{Fitness},
			grid,
			grid style=dashed,
			ticklabel style={scale=1.5},
			label style={scale=1.5},
			legend style={font=\fontsize{8}{0}\selectfont}
			]

            %\addplot[
			%color=brown,
			%mark=triangle*,
			%]
			%coordinates {
			%	(1,477026)(2,1961269)(3,7857276.5)(4,11729329)(5,15856165)(6,30264820)(7,24703945)(8,32664404)
			%};
   
            %\addplot[
			%color=black,
			%mark=*,
			%]
			%coordinates {
            %    (1,439884.5)(2,1464475.5)(3,4787594)(4,7602447)(5,13019899.5)(6,22470466)(7,27529834)(8,37681398)
			%};

            \addplot[
			color=black,
			mark=triangle*,
			]
			coordinates {
				(1,130290341.5)(2,287816051)
			};
   
            \addplot[
			color=blue,
			mark=*,
			]
			coordinates {
				(1,108291006)(2,311898164.5)
			};
   
			\addplot[
			color=red,
			mark=triangle*,
			]
			coordinates {
				(1,587173.5)(2,1236369.5)(3,2388064.5)(4,3903832.5)(5,9563556)(6,15939334.5)
			};
   
			\addplot[
			color=violet,
			mark=*,
			]
			coordinates {
				(1,947494.5)(2,1590117.5)(3,2401516)(4,5904031)(5,8615570)(6,15855735)
			};

			\end{axis}
			\end{tikzpicture}
		}
		\caption{\textit{HalfOnHalf}$\bm{R=1}$}
		\label{fig:leading:halfOnhalf:1}
	\end{subfigure}
    \begin{subfigure}[b]{0.32\linewidth}
		\resizebox{0.8\linewidth}{!}{%
			\tikzset{every mark/.append style={scale=2.5}}
			\begin{tikzpicture}
			\begin{axis}[%
            legend entries={LT-GOMEA-DirLink, LT-GOMEA-DLED, LT-GOMEA},
            legend columns=-1,
            legend to name=named,
			%legend columns=-1,
            %legend entries={dgGA,cGOMEA,P3,P3-DLED,LT-GOMEA-DLED},
            %legend to name=named,
            %legend columns=-1,
            %legend entries={LT-GOMEA-DLED,LT-GOMEA-SLL,P3-DLED,P3-SLL},
            %legend to name=named,
			xtick={1,2,3,4,5,6},
			xticklabels={50,70,100,150,200,300},
			xmin=0.5,
			xmax=6.5,
			ymode=log,
			%ymin=1e4,
			%ymax=1e9,
			%legend pos=south east,
			%xlabel=\textbf{unitation},
			ylabel=\textbf{Fitness},
			grid,
			grid style=dashed,
			ticklabel style={scale=1.5},
			label style={scale=1.5},
			legend style={font=\fontsize{8}{0}\selectfont}
			]

            %\addplot[
			%color=brown,
			%mark=triangle*,
			%]
			%coordinates {
			%	(1,477026)(2,1961269)(3,7857276.5)(4,11729329)(5,15856165)(6,30264820)(7,24703945)(8,32664404)
			%};
   
            %\addplot[
			%color=black,
			%mark=*,
			%]
			%coordinates {
            %    (1,439884.5)(2,1464475.5)(3,4787594)(4,7602447)(5,13019899.5)(6,22470466)(7,27529834)(8,37681398)
			%};

            \addplot[
			color=black,
			mark=triangle*,
			]
			coordinates {

			};
   
            \addplot[
			color=blue,
			mark=*,
			]
			coordinates {
			};
   
			\addplot[
			color=red,
			mark=triangle*,
			]
			coordinates {
				(1,587173.5)(2,1236369.5)(3,2388064.5)(4,3903832.5)(5,9563556)(6,15939334.5)
			};
   
			\addplot[
			color=violet,
			mark=*,
			]
			coordinates {
				(1,947494.5)(2,1590117.5)(3,2401516)(4,5904031)(5,8615570)(6,15855735)
			};

			\end{axis}
			\end{tikzpicture}
		}
		\caption{\textit{Alter} $\bm{R=1}$}
		\label{fig:leading:alter:1}
	\end{subfigure} 
 
%-----------------SECOND ROW-----------------------------
 \begin{subfigure}[b]{0.32\linewidth}
		\resizebox{0.8\linewidth}{!}{%
			\tikzset{every mark/.append style={scale=2.5}}
			\begin{tikzpicture}
			\begin{axis}[%
            legend entries={LT-GOMEA-DirLink, LT-GOMEA-DLED, LT-GOMEA},
            legend columns=-1,
            legend to name=named,
			%legend columns=-1,
            %legend entries={dgGA,cGOMEA,P3,P3-DLED,LT-GOMEA-DLED},
            %legend to name=named,
            %legend columns=-1,
            %legend entries={LT-GOMEA-DLED,LT-GOMEA-SLL,P3-DLED,P3-SLL},
            %legend to name=named,
			xtick={1,2,3,4,5,6},
			xticklabels={50,70,100,150,200,300},
			xmin=0.5,
			xmax=6.5,
			ymode=log,
			%ymin=1e4,
			%ymax=1e9,
			%legend pos=south east,
			%xlabel=\textbf{unitation},
			ylabel=\textbf{Fitness},
			grid,
			grid style=dashed,
			ticklabel style={scale=1.5},
			label style={scale=1.5},
			legend style={font=\fontsize{8}{0}\selectfont}
			]

            %\addplot[
			%color=brown,
			%mark=triangle*,
			%]
			%coordinates {
			%	(1,362462)(2,1171012.5)(3,2328971.5)(4,4646245)(5,14481699)(6,19300700.5)(7,24161548.5)
			%};
   
            %\addplot[
			%color=black,
			%mark=*,
			%]
			%coordinates {
            %    (1,90293.5)(2,208133.5)(3,448804)(4,1269103.5)(5,1897018)(6,2637116)(7,3650904.5)
			%};

            \addplot[
			color=black,
			mark=triangle*,
			]
			coordinates {
				(1,1584623.5)(2,2490629)(3,6780634.5)(4,13374790.5)(5,25774757)(6,95702191)
			};
   
            \addplot[
			color=blue,
			mark=*,
			]
			coordinates {
				(1,1318616.5)(2,1974687)(3,5819853)(4,16423857.5)(5,34354151)(6,109135694)
			};
   
			 \addplot[
			color=red,
			mark=triangle*,
			]
			coordinates {
				(1,587173.5)(2,1236369.5)(3,2388064.5)(4,3903832.5)(5,9563556)(6,15939334.5)
			};
   
			\addplot[
			color=violet,
			mark=*,
			]
			coordinates {
				(1,947494.5)(2,1590117.5)(3,2401516)(4,5904031)(5,8615570)(6,15855735)
			};

			\end{axis}
			\end{tikzpicture}
		}
		\caption{\textit{RestOff} $\bm{R=3}$}
		\label{fig:leading:floating:3}
	\end{subfigure}
    \begin{subfigure}[b]{0.32\linewidth}
		\resizebox{0.8\linewidth}{!}{%
			\tikzset{every mark/.append style={scale=2.5}}
			\begin{tikzpicture}
			\begin{axis}[%
            legend entries={LT-GOMEA-DirLink, LT-GOMEA-DLED, LT-GOMEA},
            legend columns=-1,
            legend to name=named,
			%legend columns=-1,
            %legend entries={dgGA,cGOMEA,P3,P3-DLED,LT-GOMEA-DLED},
            %legend to name=named,
            %legend columns=-1,
            %legend entries={LT-GOMEA-DLED,LT-GOMEA-SLL,P3-DLED,P3-SLL},
            %legend to name=named,
			xtick={1,2,3,4,5,6},
			xticklabels={50,70,100,150,200,300},
			xmin=0.5,
			xmax=6.5,
			ymode=log,
			%ymin=1e4,
			%ymax=1e9,
			%legend pos=south east,
			%xlabel=\textbf{unitation},
			ylabel=\textbf{Fitness},
			grid,
			grid style=dashed,
			ticklabel style={scale=1.5},
			label style={scale=1.5},
			legend style={font=\fontsize{8}{0}\selectfont}
			]

            %\addplot[
			%color=brown,
			%mark=triangle*,
			%]
			%coordinates {
			%	(1,477026)(2,1961269)(3,7857276.5)(4,11729329)(5,15856165)(6,30264820)(7,24703945)(8,32664404)
			%};
   
            %\addplot[
			%color=black,
			%mark=*,
			%]
			%coordinates {
            %    (1,439884.5)(2,1464475.5)(3,4787594)(4,7602447)(5,13019899.5)(6,22470466)(7,27529834)(8,37681398)
			%};

            \addplot[
			color=black,
			mark=triangle*,
			]
			coordinates {
				(1,5110722.5)(2,15809772)(3,60981297)(4,125574801)(5,428212230.5)
			};
   
            \addplot[
			color=blue,
			mark=*,
			]
			coordinates {
				(1,9663701)(2,14372963)(3,43645288)(4,182360341)
			};
   
			 \addplot[
			color=red,
			mark=triangle*,
			]
			coordinates {
				(1,587173.5)(2,1236369.5)(3,2388064.5)(4,3903832.5)(5,9563556)(6,15939334.5)
			};
   
			\addplot[
			color=violet,
			mark=*,
			]
			coordinates {
				(1,947494.5)(2,1590117.5)(3,2401516)(4,5904031)(5,8615570)(6,15855735)
			};

			\end{axis}
			\end{tikzpicture}
		}
		\caption{\textit{HalfOnHalf} $\bm{R=3}$}
		\label{fig:leading:halfOnhalf:3}
	\end{subfigure}
    \begin{subfigure}[b]{0.32\linewidth}
		\resizebox{0.8\linewidth}{!}{%
			\tikzset{every mark/.append style={scale=2.5}}
			\begin{tikzpicture}
			\begin{axis}[%
            legend entries={LT-GOMEA, LT-GOMEA-DLED, LT-GOMEA (Conc.), LT-GOMEA-DLED (Conc.)},
            legend columns=-1,
            legend to name=named,
			%legend columns=-1,
            %legend entries={dgGA,cGOMEA,P3,P3-DLED,LT-GOMEA-DLED},
            %legend to name=named,
            %legend columns=-1,
            %legend entries={LT-GOMEA-DLED,LT-GOMEA-SLL,P3-DLED,P3-SLL},
            %legend to name=named,
			xtick={1,2,3,4,5,6},
			xticklabels={50,70,100,150,200,300},
			xmin=0.5,
			xmax=6.5,
			ymode=log,
			%ymin=1e4,
			%ymax=1e9,
			%legend pos=south east,
			%xlabel=\textbf{unitation},
			ylabel=\textbf{Fitness},
			grid,
			grid style=dashed,
			ticklabel style={scale=1.5},
			label style={scale=1.5},
			legend style={font=\fontsize{8}{0}\selectfont}
			]

            %\addplot[
			%color=brown,
			%mark=triangle*,
			%]
			%coordinates {
			%	(1,477026)(2,1961269)(3,7857276.5)(4,11729329)(5,15856165)(6,30264820)(7,24703945)(8,32664404)
			%};
   
            %\addplot[
			%color=black,
			%mark=*,
			%]
			%coordinates {
            %    (1,439884.5)(2,1464475.5)(3,4787594)(4,7602447)(5,13019899.5)(6,22470466)(7,27529834)(8,37681398)
			%};

            \addplot[
			color=black,
			mark=triangle*,
			]
			coordinates {
				(1,20583460.5)(2,91026880)(3,283492972)
			};
   
            \addplot[
			color=blue,
			mark=*,
			]
			coordinates {
				(1,19148718.5)(2,95384476)
			};
   
			 \addplot[
			color=red,
			mark=triangle*,
			]
			coordinates {
				(1,587173.5)(2,1236369.5)(3,2388064.5)(4,3903832.5)(5,9563556)(6,15939334.5)
			};
   
			\addplot[
			color=violet,
			mark=*,
			]
			coordinates {
				(1,947494.5)(2,1590117.5)(3,2401516)(4,5904031)(5,8615570)(6,15855735)
			};

			\end{axis}
			\end{tikzpicture}
		}
		\caption{\textit{Alter} $\bm{R=3}$}
		\label{fig:leading:alter:3}
	\end{subfigure}
	
    \hspace{0 cm}
	\pgfplotslegendfromname{named}
	%\ref{named2}

	\caption{Median FFE until finding the optimal solution by LT-GOMEA and LT-GOMEA-DLED in solving various LBP (using $\bm{bimTrap_{10}}$) and the concatenation of $\bm{bimTrap_{10}}$ functions (X axis - problem size)}
	\label{fig:scalability}
\end{figure*}

\begin{figure}[]
    \begin{subfigure}[b]{0.49\linewidth}
		\resizebox{\linewidth}{!}{%
			\tikzset{every mark/.append style={scale=2.5}}
			\begin{tikzpicture}
			\begin{axis}[%
            legend entries={non-linearity, DLED, 2DLED},
            legend columns=-1,
            legend to name=named,
			%legend columns=-1,
            %legend entries={dgGA,cGOMEA,P3,P3-DLED,LT-GOMEA-DLED},
            %legend to name=named,
            %legend columns=-1,
            %legend entries={LT-GOMEA-DLED,LT-GOMEA-SLL,P3-DLED,P3-SLL},
            %legend to name=named,
			%xtick={1,2,3,4,5,6,7,8,9,10,11,12},
			%xticklabels={1.0, ,1.2, ,1.4, ,1.6, , 1.8, , 2.0, 2.5},
			xmin=1,
			xmax=15,
			%ymode=log,
			%ymin=0,
			%ymax=1,
			%legend pos=south east,
			xlabel=\textbf{$R$},
			ylabel=\textbf{FFE until optimal solution},
			grid,
			grid style=dashed,
			ticklabel style={scale=1.5},
			label style={scale=1.5},
			legend style={font=\fontsize{8}{0}\selectfont}
			]

            %\addplot[
			%color=brown,
			%mark=triangle*,
			%]
			%coordinates {
			%	(1,477026)(2,1961269)(3,7857276.5)(4,11729329)(5,15856165)(6,30264820)(7,24703945)(8,32664404)
			%};
   
            %\addplot[
			%color=black,
			%mark=*,
			%]
			%coordinates {
            %    (1,439884.5)(2,1464475.5)(3,4787594)(4,7602447)(5,13019899.5)(6,22470466)(7,27529834)(8,37681398)
			%};

            \addplot[
			color=red,
            name path=2dledMax,
            forget plot
			]
			coordinates {
				(1,137551652)(2,41464063)(3,28789921)(4,21132872)(5,16944112)(6,20834953)(7,15467808)(8,17008027)(9,14583237)(10,8703330)(11,9306419)(12,8222819)(13,7737999)(14,7557202)(15,6890090)

			};

            \addplot[
			color=red,
			mark=square*,
            name path=2dledMed
			]
			coordinates {
				(1,35265198.5)(2,19487320)(3,13945317.5)(4,11560300)(5,10149472.5)(6,9758759.5)(7,9178858)(8,8668876.5)(9,8486707)(10,6968153)(11,6633490.5)(12,6569469)(13,4128037.5)(14,3779173.5)(15,3894331.5)

			};

            \addplot[
			color=red,
            name path=2dledMin,
            forget plot
			]
			coordinates {
				(1,30069326)(2,9223275)(3,5651913)(4,10593978)(5,9253378)(6,7982260)(7,5252488)(8,4926531)(9,4423421)(10,3720290)(11,3501499)(12,3456873)(13,3315313)(14,3333700)(15,3234676)
			};

            \tikzfillbetween[of=2dledMin and 2dledMax]{violet, opacity=0.3};

			\end{axis}
			\end{tikzpicture}
		}
		\caption{LT-GOMEA}
		\label{fig:rscala:ltgom}
	\end{subfigure}
    \begin{subfigure}[b]{0.49\linewidth}
		\resizebox{\linewidth}{!}{%
			\tikzset{every mark/.append style={scale=2.5}}
			\begin{tikzpicture}
			\begin{axis}[%
            legend entries={non-linear, DLED, 2DLED},
            legend columns=-1,
            legend to name=named,
			%legend columns=-1,
            %legend entries={dgGA,cGOMEA,P3,P3-DLED,LT-GOMEA-DLED},
            %legend to name=named,
            %legend columns=-1,
            %legend entries={LT-GOMEA-DLED,LT-GOMEA-SLL,P3-DLED,P3-SLL},
            %legend to name=named,
			%xtick={1,2,3,4,5,6,7,8,9,10,11,12},
			%xticklabels={1.0, ,1.2, ,1.4, ,1.6, , 1.8, , 2.0, 2.5},
			xmin=1,
			xmax=15,
			%ymode=log,
			%ymin=0,
			%ymax=1,
			%legend pos=south east,
			xlabel=\textbf{$R$},
			ylabel=\textbf{FFE until optimal solution},
			grid,
			grid style=dashed,
			ticklabel style={scale=1.5},
			label style={scale=1.5},
			legend style={font=\fontsize{8}{0}\selectfont}
			]

            %\addplot[
			%color=brown,
			%mark=triangle*,
			%]
			%coordinates {
			%	(1,477026)(2,1961269)(3,7857276.5)(4,11729329)(5,15856165)(6,30264820)(7,24703945)(8,32664404)
			%};
   
            %\addplot[
			%color=black,
			%mark=*,
			%]
			%coordinates {
            %    (1,439884.5)(2,1464475.5)(3,4787594)(4,7602447)(5,13019899.5)(6,22470466)(7,27529834)(8,37681398)
			%};

            \addplot[
			color=blue,
            name path=nonLinMax,
            forget plot
			]
			coordinates {
				(1,159771925)(2,87103812)(3,29350208)(4,18457934)(5,15621191)(6,10370562)(7,11143120)(8,16252108)(9,15710712)(10,12827954)(11,12110531)(12,12242697)(13,11813916)(14,13323530)(15,11358435)
			};
   
            \addplot[
			color=blue,
			mark=*,
            name path=nonLinMed
			]
			coordinates {
				(1,55617107.5)(2,27222232)(3,15564811)(4,11975101.5)(5,9549486)(6,8340699.5)(7,7460222.5)(8,7317324)(9,7430636.5)(10,6440595)(11,6322487.5)(12,6292309.5)(13,5956384)(14,5653992)(15,5535102)
			};

            \addplot[
			color=blue,
            name path=nonLinMin,
            forget plot
			]
			coordinates {
				(1,17130104)(2,15931979)(3,9675571)(4,6629519)(5,6004272)(6,3820849)(7,4948089)(8,4490032)(9,4080604)(10,3969303)(11,3876769)(12,3644076)(13,3693120)(14,3554956)(15,2147669)
			};

            \tikzfillbetween[of=nonLinMin and nonLinMax]{blue, opacity=0.3};

			\end{axis}
			\end{tikzpicture}
		}
		\caption{LT-GOMEA-DLED}
		\label{fig:rscala:ltgomDLED}
	\end{subfigure}

	%\hspace{10.11 cm}
	%\ref{named}
	%\ref{named2}

	\caption{$\bm{R}$-based scalability: min, max, and med FFE necessary for finding the optimal solution (20 runs considered) to 150-bit Leading Block Problem \textit{RestOff} using $\bm{bimTrap_{10}}$}
	\label{fig:rscala}
\end{figure}

The hop-based statistics for the considered benchmarks are presented in Table \ref{tab:bench:hopNature}. We report the average and maximum number of hops per modification ($Hops_{max}$). Three problems on the left side of the table are concatenations of deceptive functions. Thus, a low number of hops is expected -- the average hop number per modification is close to one. Similarly, $Hops_{max}$ is also low for such problems -- its median value for all runs is two, and the maximum is three (except for the 1600-bit bimodal noised function, for which it is four). The blocks of dependent genes heavily overlap for the three problems on the right side. Nevertheless, even for a 1521-bit instance of ISG (for which the median improvements number is almost 400), the number of hops per modification is close to one, and the maximum $Hops_{max}$ is four.\par

The above results show the fundamental difference between WP\_LFL and the hard optimization problems frequently considered in GA-related research. In WP\_LFL, we need to introduce improvements one by one. In problems like Max3Sat or ISG, many improvements are independent. Oppositely, for LBP instances, a given block can be optimized only if the preceding blocks are optimal.
\par

\section{Experiments on Leading Blocks Problem}
\label{sec:leadingBlocks}

In this section, we investigate if the proposed LBP versions are harder to solve for the state-of-the-art optimizers than the concatenations of deceptive functions. We consider LT-GOMEA using SLL and LT-GOMEA-DLED using ELL. The computation budget is $10^9$ FFE, as we wish to ensure optimizers have enough time to converge. Each experiment was repeated 30 times. We consider all types of LBP with $R=1$ and $R=3$. We use $bimTrap_{10}$ as a subfunction and $\alpha = 0.1$. To recall, the set of modifications that improve the subsequent blocks in LBP shall also lead to improvements in $bimTrap_{10}$ concatenation. The difference is that in the latter case, we can improve the blocks in any order we like. Therefore, we compare the median FFE necessary to find the optimal solution to LBP and to the corresponding $bimTrap_{10}$ concatenation. \par

As presented in Figure \ref{fig:scalability}, the scalability of LT-GOMEA and LT-GOMEA-DLED is almost the same for all problems. However, for LBP instances, they scale significantly worse than for the concatenation of $bimTrap_{10}$. These two observations lead to the conclusion that the main reason for their poor performance for LBP is the lack of information on which block should be optimized first. This is also supported by the fact that for all considered LBP types, the performance of both LT-GOMEA versions is significantly higher for $R=3$. \par

As expected, the \textit{RestOff} version is the easiest to solve. Both LT-GOMEA versions scale significantly better for the  \textit{HalfOnHalf} than for the \textit{Alter} version. The \textit{HalfOnHalf} version introduces additional variable dependencies that arise from $f_d(\vec{x})$, which may make the identification of enabled blocks more challenging. Oppositely, in the \textit{Alter} version, the dependencies arising from $f_d(\vec{x})$ are the same as for the enabled blocks. Thus, it seems that for the \textit{Alter} version, the order of the optimized blocks is more important than in the \textit{HalfOnHalf} case.\par

The conclusion that the order in which blocks must be optimized can significantly influence the optimizer's effectiveness is confirmed by the results presented in Figure \ref{fig:rscala}. We present the FFE ranges necessary for finding the optimal solution. The increasing value of $R$ makes the order of block optimization less relevant. For most of the considered $R$ values, its influence on the optimizer's effectiveness is negligible. However, when $R$ drops below $3$, the FFE necessary to find the optimal solution rapidly increases. Surprisingly, the minimal FFE value remains relatively low also for the low values of $R$. We interpret this result as follows. If an optimizer is "lucky" to randomly choose the appropriate order of block optimization, then the difficulty of this process remains on a level similar to high values of $R$.\par

The above analysis shows that to solve LBP more effectively, we need to find new techniques of problem structure decomposition. The requested techniques should identify the blocks, but they must also indicate the directional dependency between them, i.e., we need information on which block should be optimized first. Finally, let us assume that we propose a linkage learning technique that improves the effectiveness of various optimizers for LBP. In that case, this technique should also lead to finding results of significantly higher quality for the WP\_LFL problem.\par

\section{Conclusions}

This paper proposes the hop-based analysis of the WP\_LFL problem instances. Its results indicate that the structure of WP\_LFL may resemble some of the features of the Leading Ones Problem. Therefore, we propose LBP, a more general version of Leading Ones and other benchmarks referring to Leading Ones that we have found in the literature. The main advantage of the proposed LBP is that it allows of the construction of many hard-to-solve problems depending on the subfunction choice and the choice of the LBP version.\par

Leading Ones and its versions are frequently a subject of the theoretical analysis. In this paper, we show that their features may be shared by some real-world problems. Finally, the results show that the problem decomposition techniques that can model the order of the block-wise dependencies are needed to solve WP\_LFL and LBP effectively. This research direction together with the hop-based analysis of other real-world problems will be the main subject of the future work.

\begin{acks}
This work was supported by the Polish National Science Centre (NCN) under Grant 2022/45/B/ST6/04150.
%and the statutory funds of the Department of Systems and Computer Networks, Wroclaw University of Science and Technology.
\end{acks}

	   \bibliographystyle{ACM-Reference-Format}
	   \bibliography{hopLeadingBlocks} 
	
\end{document}